\documentclass[10pt,journal,compsoc]{IEEEtran}

\usepackage{cite}
\usepackage{amsmath,amssymb,amsfonts, amsthm}
\usepackage{graphicx}
\graphicspath{{images/}}
\usepackage{textcomp}
\usepackage{caption}
\usepackage{subcaption}
\usepackage{xcolor}
\usepackage{hyperref}
\usepackage{tabularx}
\usepackage{longtable}
\usepackage{tabu}
\usepackage{multirow}
\usepackage{enumitem}
\usepackage{algorithm}
\usepackage{algpseudocode}
\usepackage{stackengine}
\usepackage{multirow}
\usepackage{multicol}
\usepackage{float}
\usepackage[flushleft]{threeparttable}
\usepackage{verbatim}

\begin{document}
%
\title{A Federated Approach for Fine-Grained Classification of Fashion Apparel}
%
%
%
%

\author{Tejaswini Mallavarapu, Luke Cranfill, Junggab Son, Eun Hye Kim, Reza M. Parizi, and John Morris
\IEEEcompsocitemizethanks{

\IEEEcompsocthanksitem T. Mallavarapu, L. Cranfill, J. Son, E.H. Kim, are with the Department of Computer Science, Kennesaw State University, Marietta, GA 30060. \protect\\
E-mail: see http://i2s.kennesaw.edu
\IEEEcompsocthanksitem R.M. Parizi is with the Department of Software Engineering and Game Development, Kennesaw State University, Marietta, GA 30060.
\protect\\
E-mail: rparizi1@kennesaw.edu
\IEEEcompsocthanksitem J. Morris is with Oracle Retail Global Business Unit, Atlanta, GA, USA. \protect\\
E-mail: john.x.morris@oracle.com
}
\thanks{This work was supported by the Oracle Retail Applied Research under Grant KSFP30CCSEDD.}}

\IEEEtitleabstractindextext{%
\begin{abstract}
As online retail services proliferate and are pervasive in modern lives, applications for classifying fashion apparel features from image data are becoming more indispensable. Online retailers, from leading companies to start-ups, can leverage such applications in order to increase profit margin and enhance the consumer experience. Many notable schemes have been proposed to classify fashion items, however, the majority of which focused upon classifying basic-level categories, such as T-shirts, pants, skirts, shoes, bags, and so forth.
In contrast to most prior efforts, this paper aims to enable an in-depth classification of fashion item attributes within the same category. Beginning with a single dress, we seek to classify the type of dress hem, the hem length, and the sleeve length. The proposed scheme is comprised of three major stages: (a) localization of a target item from an input image using semantic segmentation, (b) detection of human key points (e.g., point of shoulder) using a pre-trained CNN and a bounding box, and (c) three phases to classify the attributes using a combination of algorithmic approaches and deep neural networks. The experimental results demonstrate that the proposed scheme is highly effective, with all categories having average precision of above 93.02\%, and outperforms existing Convolutional Neural Networks (CNNs)-based schemes.
\end{abstract}

}

\maketitle

\IEEEdisplaynontitleabstractindextext

%
\IEEEpeerreviewmaketitle

\IEEEraisesectionheading{\section{Introduction}\label{sec:introduction}}

An application for classifying fashion apparel features in images (e.g., sleeve length, hem length, skirt style, and so forth) can be used by online retailers to help achieve a variety of goals. The merit of this proposed application arises from the fact that the annotation or attributing of fashion apparel has not kept pace with the retailer’s ability to effectively use detailed item attributes. Detailed attributes, beyond style and color, can be used for such tasks as price optimization and similar item searches. The former can contribute to increased profit margin and the latter can contribute to an enhanced consumer experience. In any case, the key insight is that a host of useful features are embedded in the retailer’s fashion apparel image assets. If one had a means of extracting those features, then one could use these extracted features to annotate fashion apparel items with a consistent and coherent set of detailed attributes. Ultimately, the case can be made that an application for classifying fashion apparel features in a retailer’s image assets would contribute substantially to closing the gap between the retailer’s aspirations and the current reality with regard to detailed fashion apparel attributes.


Several schemes have been proposed to classify fashion items. Liu \textit{et al.} proposed a human detector based on multiple features extracted by Histograms of oriented gradients, local binary patterns, and color~\cite{Liu2012}.  Eshwar \textit{et al.} proposed an apparel classification using Convolutional Neural Networks (CNN)~\cite{Eshwar2016}. Their scheme could classify five classes of T-shirt, pants, footwear, saree, and women kurta, each of which has a great discrepancy among other classes. Iliukovich-Strakovskaia \textit{et al.} proposed a fine-grained apparel categorization system with two-level classes~\cite{IS2016}.  It has 10 super-classes, such as high heels and slip-on shoes, each of which containing 14 classes inside. Seo \textit{et al.} proposed a fashion image classification by applying a hierarchical classification structure, named Hierarchical CNN (H-CNN), on apparel images~\cite{Seo2018}. They defined three levels of hierarchical classes to reduce the classification error of labeling. For every fashion item, their scheme outputs three hierarchical labels, e.g., `Clothes'-`Tops'-`T-shirt'. These results give novelty and show possibilities in fashion apparel classifications. 

Much of the extant detection, localization, and classification work focuses upon basic-level categories, which for people provide maximum information with the least cognitive effort. Specifically, one replaces a host of features with a symbol (e.g., “dog” or “bird”) that encompasses most, but not all features one might encounter in a given situation.  Such classification work is inherently built on the notion that people perceive the world around them as structured information rather than as arbitrary or unpredictable attributes. Our work, in contrast to most prior efforts, seeks not to distinguish between categories, but to distinguish between instances of the same category in a meaningful and consistent manner. Dresses, for example, have numerous subordinate features depending on their length~\cite{Garments}. Sleeve lengths categorized them into a cap, short, elbow length, bracelet, long, and angel. Hem lengths categorized them into mini, above-the-knee, knee, below-the-knee, mid-calf, evening, and floor. 

Moreover, we do not begin our analysis of an image with vague generalities concerning its content.  Rather, we begin our analysis of an image with numerous priors. Specifically, we know a priori what the image we are analyzing depicts, e.g., a model wearing a dress against a neutral background—prior use of images in catalogs and on websites ensures that basic category information is available.  Given that we know the instance category, we also know what to look for next, e.g., hem lengths, sleeve lengths, necklines, and so on.  Thus, our problem begins with localizing a known target category and then classifying the localized category instance with respect to one or more features known to be present, such as hem length for dresses.  

On the basis of these observations, we propose a federated approach that consists of both deep neural networks and an algorithmic approach. Specifically, our proposed scheme is constructed using SegNet~\cite{badrinarayanan2015segnet} to localize target categories and an algorithmic approach to classify localized category instances. The algorithmic approach leverages human key-point detection~\cite{cao2020realtime} and bounding box schemes in order to improve the classification performance. We analyzed the performance of our model on classifying attributes of localized categories such as hem length, sleeve length and hem type against three CNN variant models using metrics like precision, recall, and f1 score. The average f1 score of the hem length category of our model is 97.12\%  while CNN, VGG16, and VGG19 networks have 49.1\%, 50.5\%, and 50.23\%, respectively. For the sleeve length category, the average f1 score of our model is 89.45\% whereas for CNN, VGG16, and VGG19 have 76.95\%, 49.62\%, and 46.2\%, respectively. Similarly, for the hem type category, our model achieved the highest f1 score of 85.04\%  while CNN, VGG16, and VGG19 have 80.12\%, 69.26\%, and 76.36\%, respectively. These experimental results demonstrate that our model outperformed the other models in all categories. 


The remainder of this paper is organized as follows. Section \ref{sec:relatedWork} gives the related work. Section \ref{sec:background} provides  essential backgrounds and building blocks underpinning the research. The proposed scheme is presented in Section \ref{sec:proposed}. Section \ref{sec:experiments} gives the evaluation results with respect to the effectiveness of the proposed scheme . Finally, Section \ref{sec:conclusion} concludes this paper.

\begin{figure*}[h]
		\centering
		\centerline{\includegraphics[width=1\linewidth]{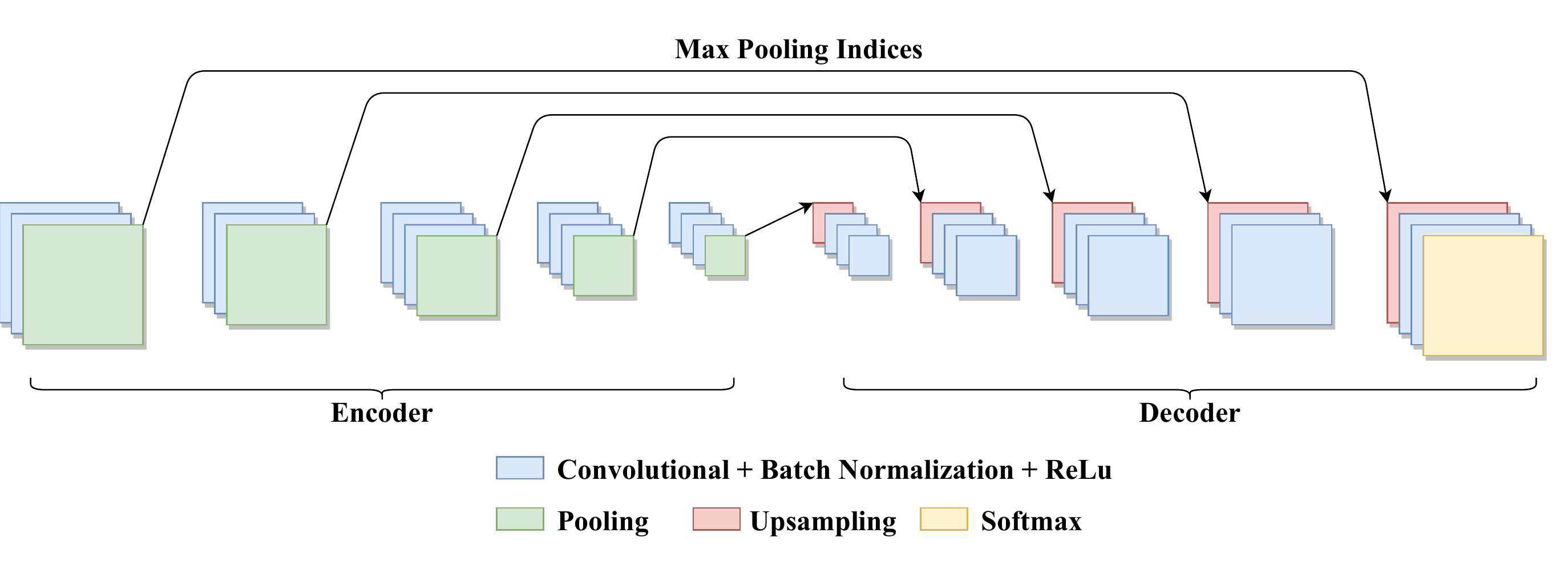}}
		\caption{SegNet Architecture}
		\label{figure1}
\end{figure*}

\section{Related Work}\label{sec:relatedWork}
Image classification is a supervised learning problem that aims to identify objects or scenes in images. The advent of machine learning has helped propel this field forward, with the creation of CNN being the leading factor.  
This section introduces related works in fashion item detection literature, and number of key computer vision components. 


\subsection{Semantic Segmentation}
Fashion item classification is an area of research that uses image segmentation to great effect. Although many schemes have been developed and utilized, Fully Convolutional Network (FCN) is one of the most popular algorithms ~\cite{Shelhamer2014FullyCN}. As the name suggests, the model is fully convolutional, containing no fully connected layers. This allows the algorithm to receive image input of any size and circumvents difficulties such as pre- and post-processing complications. 

A deep convolutional Encoder-Decoder Architecture for image segmentation (SegNet) is worthy of remark~\cite{Badrinarayanan2017SegNetAD, Hu2008ClothingSU}. SegNet consists of an encoder, decoder network, and pixel-wise classification layer. The encoder network corresponds to the 13 convolutional layers in the VGG16 network introduced in ~\cite{simonyan2014deep}. It performs convolution with a filter bank to produce a set of feature maps. 

Other deep learning architectures have also been introduced for semantic segmentation, including U-Net~\cite{Ronneberger2015UNetCN}, Region based Convolutional Neural Network (R-CNN) families~\cite{R-CNN, FastR-CNN, FasterR-CNN, He2015DeepRL, He2017MaskR}, and YOLO~\cite{Redmon2018YOLOv3AI, Howard2017MobileNetsEC}. 

U-Net is similar to FCNs, however, it is modified, which gives better segmentation in medical images~\cite{Ronneberger2015UNetCN}. The main differences when compared to FCNs are: U-Net is symmetric, and the skip connections apply a concatenation operator between the downsampling path and the up-sampling path. One of the main advantages of U-Net is that it is much faster than FCN and MASK R-CNN~\cite{He2017MaskR}, with segmentation of a 512x512 image taking less than 1 second on a GPU.

Faster R-CNN is an algorithm that is typically used for object detection~\cite{He2017MaskR}. It contains two steps: a Region Proposal Network (RPN) that generates proposals for object bounding boxes and Fast R-CNN to extract features using RoIPool from the object bounding boxes and perform classification~\cite{Ronneberger2015UNetCN}. Mask RCNN is an extension from Faster RCNN, in which a third branch that gives the binary object mask. However, since the much finer spatial layout of input is required in additional mask output, Mask RCNN uses the FCN, and Mask RCNN results in more accurate outputs than FCN. 

\subsection{Pose Estimation}
Pose estimation or landmark detection focuses on detecting key features on an image, e.g., eyes and nose on a human. The beginning of rapid expansion in this field can be traced back to when neural networks were first used for pose estimation in 2014 \cite{Toshev_2014_CVPR}. Following this, another influential work was published in 2015, a model to output the results as a heat map instead of a single point \cite{tompson2015efficient}. The practice of using heat-maps as output for landmark detection has shaped many different models. Many architectures in human pose estimation and particularly fashion item detection now use multi stage-architecture first developed in 2016 \cite{wei2016convolutional}. Other architectures have taken a different approach, using a self correcting model that used Iterative Error Feedback, that is, errors are corrected iteratively instead of all at once \cite{carreira2016human}. In 2016 a new model used an "hourglass" architecture that does repeated up-sampling then down-sampling, and at the time it was released it was the highest performing architecture in its field  \cite{newell2016stacked}. Pose estimation was characterized by large and computationally expensive models until an archived work attempted to make the simplest architecture possibly and performed competitively on the COCO data-set~\cite{xiao2018simple}. As of 2019, the highest performing model was the High-Resolution Network (HRNet)~\cite{sun2019deep}. As the name implies, this networks novelty lays in the fact that it keeps the image at a high resolution throughout the process. These significant advances in human pose estimation over the years have helped push forward the field of fashion item detection as well. 

\subsection{Fashion Apparel Detection }


Ye et al. proposed Finer-Net, which is a cascaded human parsing with hierarchical granularity~\cite{8486505}. Finer-Net consists of three stages, going from higher level to lower level features with each stage. The use of high level to low level features allows for a model robust to occlusion commonly seen in fashion item detection. MH-Parser is a novel multi-human parsing model and introduced with a new multi-human parsing (MHP) dataset by Jianshu Li~\cite{Li2017MultipleHumanPI} and is one of the few models in the fashion detection field to implement a Generative Adversarial Network (GAN) for its detection. Additionally, when MH-Parser performs global parsing prediction, the MH-Parser uses a fully convolutional representation learner shared for global parsing and map prediction. 

Supplementary to the FCN and equally important has been the use of human pose estimation and fashion landmarks. Extracting fashion features plays an important role when performing Fashion Apparel Detection~\cite{Liu2016FashionLD, 8410445, 6898848, Xiao2018SimpleBF, 8578547, 8953890, 6909508, 8486505}. Experimental results from various papers show that clothing categories and attributes can be classified by using fashion landmarks, key-points, and pose estimations. By utilizing these methods, the accuracy of detected fashion items can be increased~\cite{8486505, 8953890}. 

One of the most influential works in this field was the first (to the best of our knowledge) to propose fashion landmarks around things like the neckline in 2016 \cite{liu2016fashion}, additionally they released a data-set annotated with these landmarks which paved the way for many researchers to use these to improve their models. Since then researchers have been seeking to improve fashion landmark detection with different models \cite{Chen_2019_ICCV}. Among some of the best models using landmark detection for fashion item detection is a model based on assistance from high level industry knowledge in the form of grammars to guide the fashion landmarks that significantly improve performance \cite{wang2018attentive}. Additionally, it was shown in 2019 that  a model using landmark detection to bolster accuracy outperformed all state of the art methods on the Deep Fashion data-set \cite{Lee_2019_ICCV}. Researchers recently attempted to improve landmark detection, which can over generalize, by using multiple layer Layout-Graph Reasoning, with each layer mapping to another set of features, one layer for body parts, coarse clothes features, finer features, etc. and achieved impressive results, also contributing a new data-set with 200,000 images \cite{Yu_2019_CVPR}.

In an archived paper \cite{jia2018deep} some researchers took a new perspective on fashion item detection and sought to do fashion item detection for industry professionals instead of consumers. In contrast to the land-marked images some researchers \cite{8451125} sought to use weakly annotated images with multi-stage architecture to perform fashion item detection. This work was similar to our own in that it reached into some of the finer details of fashion item detection, but only used website keywords such as "sleeve" or "v-neck" instead of creating detailed categorical divisions among these finer grained attributes. Additionally, this work exclusively used a CNN while our work uses both a CNN and algorithmic approach. In the same vein, some authors \cite{Rubio_2017_ICCV} attempted to use keyword metadata that accompanied images to predict the main product in each image. For a more complete set of works in the fashion field, we direct our readers to an archived survey done this year that examines more than 200 works in the field \cite{cheng2020fashion}.

\begin{figure*}[h]
		\centering
		\centerline{\includegraphics[width=1\linewidth]{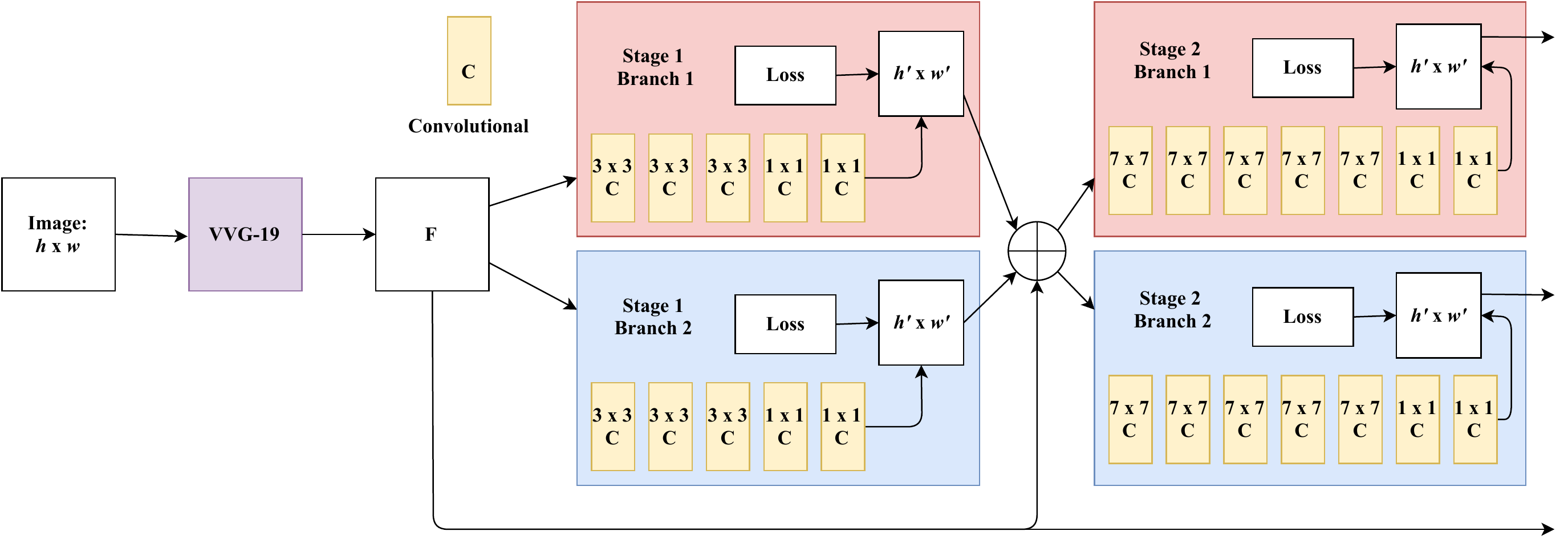}}
		\caption{Key Point Architecture}
		\label{figure2}
\end{figure*}

\section{Building Blocks}\label{sec:background}

\subsection{SegNet}
SegNet is a neural network for pixel-wise semantic segmentation, originally developed by Cambridge \cite{badrinarayanan2015segnet} in 2015. At the time of its development, its novelty laid in the encoder decoder setup for up-sampling. The SegNet encoder architecture is identical to the convolutional layers in VGG16 \cite{simonyan2014deep}, but the fully connected layers are removed. The architecture is depicted in Figure~\ref{figure1}. This means the SegNet is fully convolution, allowing it to accept any input size and significantly reducing the memory and computational power needed, making it a practical, scalable solution.  

The SegNet is effective because it saves the indices while doing max pooling that reduces the image size by taking the most prominent feature of a section and selecting it to represent that section in the reduced image. Each encoder layer in the SegNet has a corresponding identical decoder layer. The decoder upsamples its feature maps using the memorized max-pooling indices from the corresponding encoder feature map and produces sparse feature maps, which are then convolved with a kernel to produce dense feature maps. Once this up-sampling is complete, there is a soft-max classification layer for pixel-wise classification.

\subsection{Bounding Boxes}
Bounding boxes are a very common tool for object detection in the field of computer vision. Bounding boxes are rectangles defined by the upper left corner and the lower right corner on an \textit{(x, y)} coordinate plane. While neural networks can predict bounding box locations, machine learning was not used in this part of the project.

The \textit{(x, y)} coordinates are identified by pixel values, this identification of target pixels was made simple by semantic segmentation of the images. Once the indices are known, OpenCV library~\cite{opencv_library} is used to draw a rectangle around the desired portion of the image.

\subsection{Key Point Identification}
Key point identification of humans stemmed from the problem of pose estimation in computer vision. The goal of the task is to identify landmarks or key points on an object, in this case a human. In 2016 researchers at Carnegie Melon University, won the COCO Keypoints Challenge with their model~\cite{cao2016realtime, cao2020realtime}. As illustrated in Figure~\ref{figure2}, their architecture relied on very deep learning and works in three stages. 

The preliminary stage, stage 0, generates a set of feature maps of a given image, using the first 10 layers of VGG-19 \cite{simonyan2014deep}. The set of feature maps is given as input to the stage 1, next stage of the architecture. Stages 1 and 2 have two branches each. The first branch predicts confidence maps, and the second branch predicts affinity fields. The confidence maps are a heat map of confidence levels of a certain key point falling in a region. The affinity fields are vectors that show the degree of association between body parts. The confidence maps and affinity fields are concatenated with the original set of feature maps to increase prediction accuracy, and are fed into stage two. Two L2 loss functions are used at the end of each stage, one for each branch. Stage 2 then parses the affinity fields and confidence maps with greedy inference to produce the final output, which is a set of 2D points which are the coordinates for the detected key points.

\section{Proposed Scheme}\label{sec:proposed}

\begin{figure*}[h]
		\centering
		\centerline{\includegraphics[width=1\linewidth]{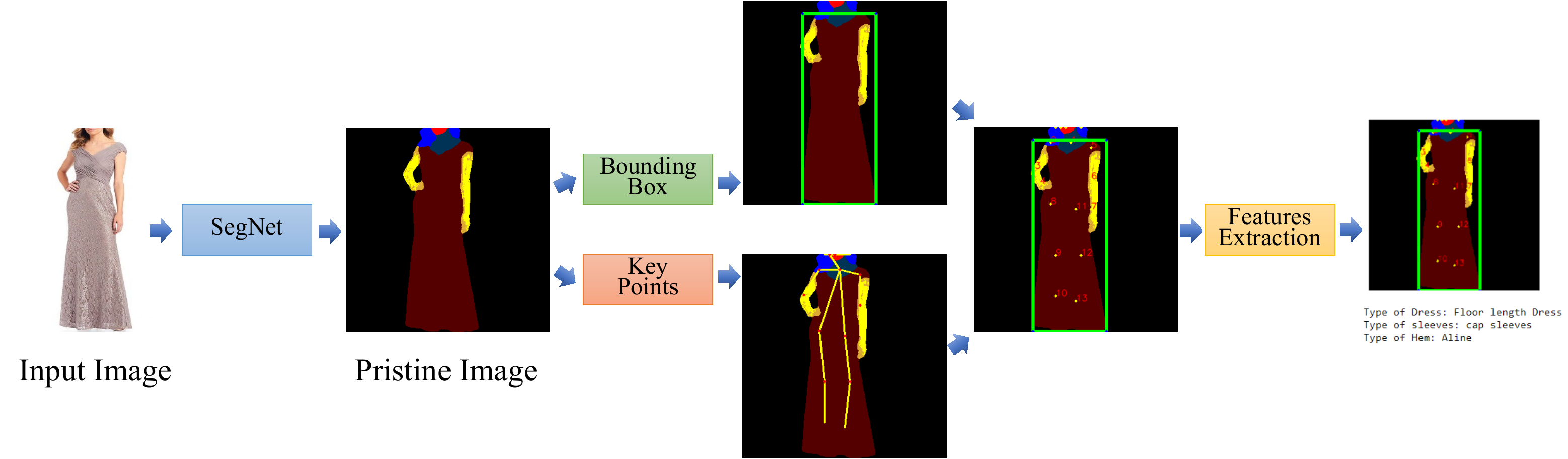}}
		\caption{An Overview of the proposed architecture}
		\label{figure:proposed}
\end{figure*}

\begin{table}
\centering
\caption{Classes and Attributes of Datasets}
\label{tab:labels}
    \begin{tabular}{ |c|c| } 
     \hline
     \textbf{Classes} & \textbf{Attributes} \\ \hline \hline
     \multirow{10}{*}{Hem Length} & Floor Length  \\ \cline{2-2}
     & Evening           \\ \cline{2-2}
     & Lower Calf       \\ \cline{2-2}
     & Below Midcalf   \\ \cline{2-2}
     & Midcalf          \\ \cline{2-2}
     & Below Knee        \\ \cline{2-2}
     & Knee            \\ \cline{2-2}
     & Above Knee      \\ \cline{2-2}
     & Mini             \\ \cline{2-2}
     & Micro              \\
     \hline
     \multirow{5}{*}{Sleeve Length} & Long    \\ \cline{2-2}
     & Bracelet \\ \cline{2-2}
     & Elbow \\ \cline{2-2}
     & Short  \\ \cline{2-2}
     & Cap  \\ 
     \hline
     \multirow{4}{*}{Hem Type}& Aline  \\ \cline{2-2}
     & Straight  \\\cline{2-2}
     & High-low  \\ \cline{2-2}
     & Asymmetrical   \\
     \hline
    \end{tabular}
\end{table}

\subsection{Overview}
Given an image with a dress, the proposed scheme aims to classify find-grained attributes of the dress. The target attributes include "Hem Length", "Sleeve Length", and "Hem Type", which can be determined based on its length information. The "Hem Length" will be classified to one of ten attributes, such as 'Floor Length', 'Evening', 'Lower Calf', 'Midcalf', 'Below Midcalf', 'Below Knee', 'Knee', 'Above Knee', 'Mini', and 'Micro'. The "Sleeve Length" will be classified to one of five attributes, such as 'Long', 'Bracelet', 'Elbow', 'Short', and 'Cap'. Finally, the "Hem Type" will be classified to one of four attributes, such as 'Aline', 'Straight', 'high-low', and 'Asymmetrical'. Table~\ref{tab:labels} summarizes these classes and their attributes derived from~\cite{Garments}.

Fig.~\ref{figure:proposed} illustrates an overview of the proposed scheme. It consists of five phases:
\begin{itemize}
\item Stage 1: Extracts dress from the input image with semantic segmentation.
\item Stage 2: Estimates human key joints (key points) and creates bounding box over the dress region.
\item Stage 3
\begin{itemize}
    \item Phase 1: Classify the hem length.
    \item Phase 2: Classify the style of dress sleeves.
    \item Phase 3: Describes the hem styles of dress.
\end{itemize}

\end{itemize}
With all these stages, our proposed federated approach classify localized categories and generates detailed descriptions of the dress. For example, the output will be "A-line floor length dress with cap sleeves".


\subsection{Stage 1: Dress Extraction}

This stage uses SegNet model to separate the dress from noises like background and skin.
The encoder consists of 13 convolutional layers in five blocks and each convolutional layer applies batch normalization, ReLU activation and then max pooling layer with  2 x 2 filters and stride 2. The decoder also has 13 convolutional layers similar to encoder except non linearity. 
The SegNet model takes input as either a RGB color image or a gray scale image and output a segmentation map where each pixel is assigned with a class label. 
To train the model, first all the input images are resized into 320 x 320 pixels and the proper initialization of parameters are required as bad initialization can hinder the learning of the networks.
Therefore, a robust weight initialization method proposed in \cite{7410480} is used to initialize the weights of encoder and decoder networks whereas biases are initialized with zero.

The hyper-parameters used in this model are as follows. The learning rate is 0.01, momentum is 0.9 and the mini-batch size is 128 with 50,000 epochs.
All the parameters are trained with stochastic gradient descent (SGD) until the training loss converges.
The training data is shuffled for each epoch to ensure that each image is used only once in an epoch. 
The objective function of this model is the categorical cross-entropy loss which is computed on a mini-batch. 
Larger variation in the number of pixels in each class requires computing the loss differently based on true class.
To get smooth segmentation, median frequency class balancing method is applied.
This method ensures that smaller classes in the training data have higher weights whereas the larger classes have weights smaller than 1.

\subsection{Stage 2: Keypoints Estimation and bounding box}
The feature maps obtained from the stage 1 is further processed to estimate keypoints and build a bounding box. It is worth noting that only the images with pristine feature maps from phase 1 are collected and used in this phase. The images that are misclassified or having any occlusions are removed from the prediction phases. 

To get the bounding box, feature maps from the first phase are subjected to thresholding, a type of image segmentation.
In thresholding, the binary mask with simple black and white pixel values can be obtained by using the threshold value \textit{t}. Gray-scale histogram of the image gives the \textit{t} value and pixels values greater than \textit{t} will be turned ``on," while the pixel values less than \textit{t} value will be turned ``off."
Pixels that are turned on belongs to the dress region (ROI) and is covered in one color while the remaining predicted labels including background that are turned off are grouped into one color.   
Then the bounding box is constructed by computing the minimum and maximum \textit{(x, y)} coordinates of the dress region.

The human pose estimation model followed in this paper is proposed by \cite{8099626}. The architecture of the model is described in section IV.
The model is pre-trained on coco data set and generates Confidence Maps and Part Affinity maps which are all concatenated. The coco dataset has 18 keypoints including body, foot, hand, and facial keypoints.
This model takes raw input image and output the four-dimensional matrix of which the first dimension describes the image ID. The second dimension indicates the indices of a keypoints which include 18 keypoint confidence Maps, 1 background and 38 Part Affinity Maps. 
The third and fourth dimension represents the height and width of the output map, respectively.
Once the indices of keypoints are known, we locate the same indices on corresponding feature maps that are generated from phase I. The resultant output map that has both bounding box coordinates and keypoints is used as input for all the remaining phases. 

\subsection{Stage 3: Dress Attributes Classification}
In stage 3, the algorithmic approaches of our proposed scheme for classifying dress categories are described. Each phase deals with classification of one category. The output map of stage 2 is passed through each phase to get the final output with hem length, sleeve length and hem type classification.

\subsubsection{Phase 1: Hem Length Classification}
In this phase, each output image is mapped to one of the 10 hem lengths by using Algorithm~\ref{algo:Length}. The annotations of human keypoints and a bounding box helps in computing the hem length and leg length. Hem length is determined as the distance between hip keypoint to end of the dress while leg length is measured as the distance between hip to ankle keypoint. By computing the ratio ($H_\ell$) of hem length to leg length, we defined 10 hem length attributes. Threshold for each attribute is manually derived using specifications from~\cite{Garments}. The dress that has $H_\ell$ greater than or equal to 1.05 threshold value is classified as the floor length dress as it extends over the ankle, whereas the dress whose $H_\ell$ is less than 0.3 is classified as micro as it ends near high thighs and closer to hip key point. If the $H_\ell$ value is less than 1.05 and greater than or equal to 0.9 threshold, we classified it as evening dress while the dress whose $H_\ell$ is in between 0.75 and 0.9 is defined as lower calf. Likewise the dress with $H_\ell$ in between 0.675 and 0.75 falls under below midcalf and the dress with $H_\ell$ ranges from 0.6 to 0.675, 0.515 to 0.6, 0.475 to 0.515 is classified as midcalf, below knee, and knee, respectively. Above knee and mini hem lengths are defined if $H_\ell$  value ranges from 0.375 to 0.4.75 and 0.3 to 0.375.


  \begin{algorithm}[h]
    \caption{Classifying Dress Length}
    \label{algo:Length}
    \begin{algorithmic}[1]
    \Statex \textbf{Input:}
    An image with human key points and a bounding box (Output of Phase 2)
    \Statex \textbf{Output:}
    A classified result of hem length\\
    Compute $H_\ell=$ dist(hip-end of dress)/dist(hip-ankle)
    \If{$H_\ell\geq 1.05$} \State{\Return{Floor Length}}
    \ElsIf{$0.9 \leq H_\ell < 1.05$} \State{\Return{Evening} }
    \ElsIf{$0.75 \leq H_\ell < 0.9$} \State{\Return{Lower Calf} }
    \ElsIf{$0.675 \leq H_\ell < 0.75$} \State{\Return{Below Midcalf} }
    \ElsIf{$0.6 \leq H_\ell < 0.675$} \State{\Return{Midcalf} }
    \ElsIf{$0.515 \leq H_\ell < 0.6$} \State{\Return{Below knee} }
    \ElsIf{$0.475 \leq H_\ell < 0.515$} \State{\Return{Knee} }
    \ElsIf{$0.375 \leq H_\ell < 0.475$} \State{\Return{Above Knee} }
    \ElsIf{$0.3 \leq H_\ell < 0.375$} \State{\Return{Mini} }
    \ElsIf{$H_\ell < 0.3$} \State{\Return{Micro}}
    \Else \State{\Return{$\bot$}}
    \EndIf
    \end{algorithmic}
    \end{algorithm}

\subsubsection{Phase 2: Dress Sleeve Classification }
In Phase 4, each image is classified in to one of five types of sleeves using Algorithm~\ref{algo:Sleeve}. The hand keypoints of both hands are used in defining type of sleeves. There are three key points such as shoulder, elbow and wrist key points on each hand and label them as $K_S, K_E$, and $K_W$, respectively. Then define the end point of sleeve as $E_s$. Bisect the regions between $K_S $ and $ K_E$ and $K_E$ and $K_W$ and set as point $K_{SE}$ and $K_{EW}$, respectively. Then track the region where the end point of sleeve exist and classify sleeve types. The sleeve that ends in more than 5 pixels below the mid region between elbow and wrist key point ($K_{EW}$) is classified as long sleeve. The sleeve that lasts in the region of 5 pixels above and below $K_{EW}$ point is defined as bracelet sleeve while the sleeve ends in the region of 5 pixels above and below $ K_E$ key point is classified as elbow. The sleeve whose end point is in region of 5 pixels above and below $ K_{SE}$ is defined as short otherwise if find in the region near shoulder key point, it is labeled as cap sleeve.  


 \begin{algorithm}[h]
    \caption{Classifying Dress Sleeve}
    \label{algo:Sleeve}
    \begin{algorithmic}[1]
    \Statex \textbf{Input:}
    An image with human key points and a bounding box (Output of Phase 2)
    \Statex \textbf{Output:} A classified result of sleeve length\\
    Find an end point of a sleeve and set the value as $E_s$ \\
    Find a shoulder point and set the value as $K_S$ \\
    Find an elbow point and set the value as $K_E$ \\
    Find a wrist point and set the value as $K_W$ \\
    Compute a bisect point, $K_{SE}=(K_S+K_E)/2$  \\
    Compute a bisect point, $K_{EW}=(K_E+K_W)/2$ 
    \If{$E_s >K_{EW}+5$ px} \State{\Return{Long}}
    \ElsIf{$K_{EW}-5\text{ px} < E_s \leq K_{EW}+5$ px}\State{\Return{Bracelet}}
    \ElsIf{$K_E-5$ px $< E_s \leq K_E +5$ px} \State{\Return{Elbow}}
    \ElsIf{$K_{SE}-5$ px $< E_s \leq K_{SE}+5 $px } \State{\Return{Short}}
    \ElsIf{$E_s \leq K_S+5$ px }\State{\Return{Cap}}
    \Else \State{\Return{$\bot$}}
   	\EndIf
    \end{algorithmic}
    \end{algorithm}

\subsubsection{Phase 3: Dress hem style Classification}
Phase 5 classifies four different hem styles based on the Algorithm~\ref{algo:Hem}. The three leg key points on each leg such as hip, knee and ankle and a bounding box are used in defining hem styles. Locate the end point of dress from any leg key point that is closer to end of dress on both left and right legs and define them as $E_L$ and $E_R$, respectively. Set the end of dress that meets the bounding box bottom as $T$. 
Asymmetrical sleeve is defined if either the absolute distance between y coordinate of bounding box bottom $T$ and the $E_L$ or $T$ and the $E_R$ is less than or equal to 5 pixels and if the absolute difference between $T$ $ - E_L$  and $T$ $ - E_R$ is greater than or equal to 5 pixels where as high-low is defined if the absolute difference between $T$ $ - E_L$  and $T$ $ - E_R$ is less than 5 pixels. On the other hand, a-line and straight hem styles are classified based on the width of bounding box. The width can be computed by difference between $ X_2 $  and $ X_1 $ coordinates of the bounding box. If the width of bounding box is greater than 110 pixels, we considered as a-line hem otherwise straight hem style. The threshold values and metrics to classify these hem styles are derived from~\cite{Garments}.

 \begin{algorithm}[h]
    \caption{Classifying Dress Hem style}
    \label{algo:Hem}
    \begin{algorithmic}[1]
    \Statex \textbf{Input:} An image with human key points and a bounding box (Output of Phase 2)
    \Statex \textbf{Output:} A classified result of hem type\\
    Find an end point of the dress from left knee or ankle and set the value as $E_L$\\
    Find an end point of the dress from right knee or ankle and set the value as $E_R$\\
    $T=$ bottom of the bounding box  
    \If{$(|T-E_L| \leq 5$ px$)  ||  (|T-E_R|\leq 5$ px$)$ }
            \If{$|E_L-E_R|\geq 5$px} \State{\Return{Asymmetrical}}
            \Else \State{\Return High-low}
            \EndIf
    
    \ElsIf{width of bounding box $\geq 110$} \State{\Return Aline}
    \ElsIf{width of bounding box $<110$} \State{\Return Straight}
    \Else \State{\Return{$\bot$}}
    
    \EndIf
    
    \end{algorithmic}
    \end{algorithm}

\section{Experimental Results and Analysis}\label{sec:experiments}

We performed all the experiments in Python using Keras and Tensorflow libraries on GPU server with four core processors for faster computation time. In this section the types of data sets, algorithm settings and results are discussed.

\subsection{Dataset Preparation}
Image segmentation experiments using the SegNet model were carried out on 50,000 images from Look into Person (LIP) dataset~\cite{Liang2019}. Each image in the set is annotated pixel-wise with 19 semantic human part labels and 1 background label for human parsing. The training, validation, and test sets consists of 30,000, 10,000 and 10,000 images, respectively. The labels of this dataset include 6 body parts such as right and left side of arms and legs and 13 clothes categories like upper clothes, pants, dress, skirts, sunglasses, gloves, shoes, and socks, etc. Each label is annotated with different RGB color encoding. For instance, all the pixels belonging to dress are encoded as [0,0,85], all pixels of right leg are encoded as [255,134,255], etc.
  
Dress classification experiments were evaluated on dress images crawled from online retail shops including Amazon, Forever21, Shein, etc. All images were gathered under fair use doctrine. We manually labeled each image with three different attributes of dresses including hem length, sleeve type, and hem type. For each subclass around 700 images are collected. All the collected images must be a full body shot images with no occlusions. Images that have a rear view of the dress or no proper posture to detect human keypoints were removed from the dataset as they prevent the successful identification of different categories.

The SegNet model which is pre-trained on the LIP dataset, takes all colored dress images as input and produces segmented maps. The segmented images with background noise and occlusions are removed and only pristine images are selected for further analysis of our model unlike CNN and VGG which uses original colored images of pristine segmented images. 


\subsection{Algorithm settings}
We evaluate the performance of our model on each dress category and compared with several state-of-art classifiers such as Convolutional Neural Network (CNN) and VGG16 and VGG19 networks.

The CNN architecture has three convolutional layers, two fully connected layers (FC) and an output layer. Each convolutional layer has a filter size of 3 × 3, and max-pooling was performed on every 2 × 2-pixel block. Batch Normalization and Dropout are  used as regularization factors. The output is then fed to Soft-max layer for classification that helps to determine the classification probabilities used by the final classification layer. The CNN model is trained with Adam optimizer with adaptive learning rate.

Both the VGG16 and VGG19 networks are deep CNN models with five building blocks. The VGG16 has a total of 16 layers while VGG19 has 19 layers. The first two blocks in both the networks have two convolutional layers and 1 pooling layer with 64 filters in the first block and 128 filters in the second block. The third and fourth block of VGG16 network consists of 3 convolutional layers and 1 pooling layer each and the last block has 3 convolutional layers whereas the VGG19 network has  4 convolutional layers and 1 pooling layer each in the third and fourth blocks and 4 convolutional layers in the fifth block. The fifth block is followed by two fully connected layers of size 4096 nodes. The 3x3 sized filters with stride of 1 are used in all convolutional layers. The third block has 256 filters while fourth and fifth blocks have 512 filters. For all the max-pooling layers, 2x2 filter with the stride of 2 is used. Batch normalization is performed at each block for easier initialization and faster training. Dropout regularization with 0.5 probability is used to prevent over-fitting between dense connected layers. Both the networks used stochastic gradient descent optimizer with 0.9 momentum and 0.001 learning rate. 

We performed hyperparameter search for determining optimal training parameters for all the three networks CNN, VGG16 and VGG19 and selected models with the lowest loss computed on the validation sets. The optimal hyperparameters obtained are 500 epochs with batch size of 24. The nonlinear ReLU is used as an activation function in all layers but output layer. In addition to avoid memory issue with our GPU server, the original image size is resized to 220 x 200 pixels.

\subsection{Evaluation Metrics}
Our main goal is to create a model that classifies dress attributes with high computational efficiency which can be used in real applications. 
In order to evaluate the classification performance of the proposed algorithm, we computed
several metrics such as precision, recall and f1 score and compared with three variants of CNN-based models.
\textit{Precision} is defined as the percentage of correctly predicted images, while \textit{recall} is defined as the percentage of predicted images a model correctly identified.
The \textit{f1 score} is defined as the weighted harmonic mean of precision and recall.
The formal definition of precision, recall and f1 score are given below:

\begin{equation}
\label{Precision}
Precision = \frac{TP}{TP+FP}
\end{equation}

\begin{equation}
\label{Recall}
Recall = \frac{TP}{TP+FN}
\end{equation}

\begin{equation}
\label{F1-Score}
F1\text{ }Score = 2 *  \frac{Precision * Recall}{Precision + Recall}
\end{equation}

\subsection{results and discussion}

The results in Table~\ref{tab:Results_our model} illustrate the performance of our model on Hem length, sleeve length and hem type attributes. The performance scores show that hem lengths such as below knee, knee and micro have 100\% precision, recall and f1 score. High precision, recall and f1 score shows that our model has very low false positive rate and false negative rate. In terms of sleeve length, except cap sleeves, the remaining four sleeves has precision greater than 85\%. On the other hand, short sleeve has the lowest recall and f1 score. The high precision and the lowest recall and f1 score of short sleeve represent that our model wrongly classified some of short sleeves as other type of sleeves. In Hem type classification, high low has the lowest precision of 83.6\% while other hem types has precision greater than 89\% . The recall and f1 score show that asymmetrical has lowest values. Further the tabular results illustrate that the overall performance of our model is higher for hem length  and lower for sleeve length classification. 



Table~\ref{tab:Results_CNN model} summarizes the performance of CNN model. It is interesting to note that CNN performs better on sleeve length and hem type than hem length classification. Except floor length, the CNN model has poor performance on remaining hem types. The low precision and high recall scores of asymmetrical attribute report that the model wrongly classifies other hem types as asymmetrical. The performance of VGG16 and VGG19 models are shown in Table~\ref{tab:Results_VGG16 model} and Table~\ref{tab:Results_VGG19 model}, respectively. Similar to CNN, both VGG16 and VGG19 networks showed poor performance on hem length data except floor length attribute. The higher performance of CNN to VGG16 and VGG19 networks infers that the inclusion of more convolutional layers has no effect in improving the classification results for this dress attribute classification problem. This demonstrates that CNN with less complex architecture can extract attribute-sensitive features for enhancing the classification. However, the average precision, recall and f1 score of our proposed scheme is higher than CNN. This indicates that the CNN model which performed better than both the VGG networks is completely inferior to our model in all the three categories.



\begin{figure*}
    \begin{subfigure}{.5\textwidth}
        \centering
        \includegraphics[width=.8\linewidth]{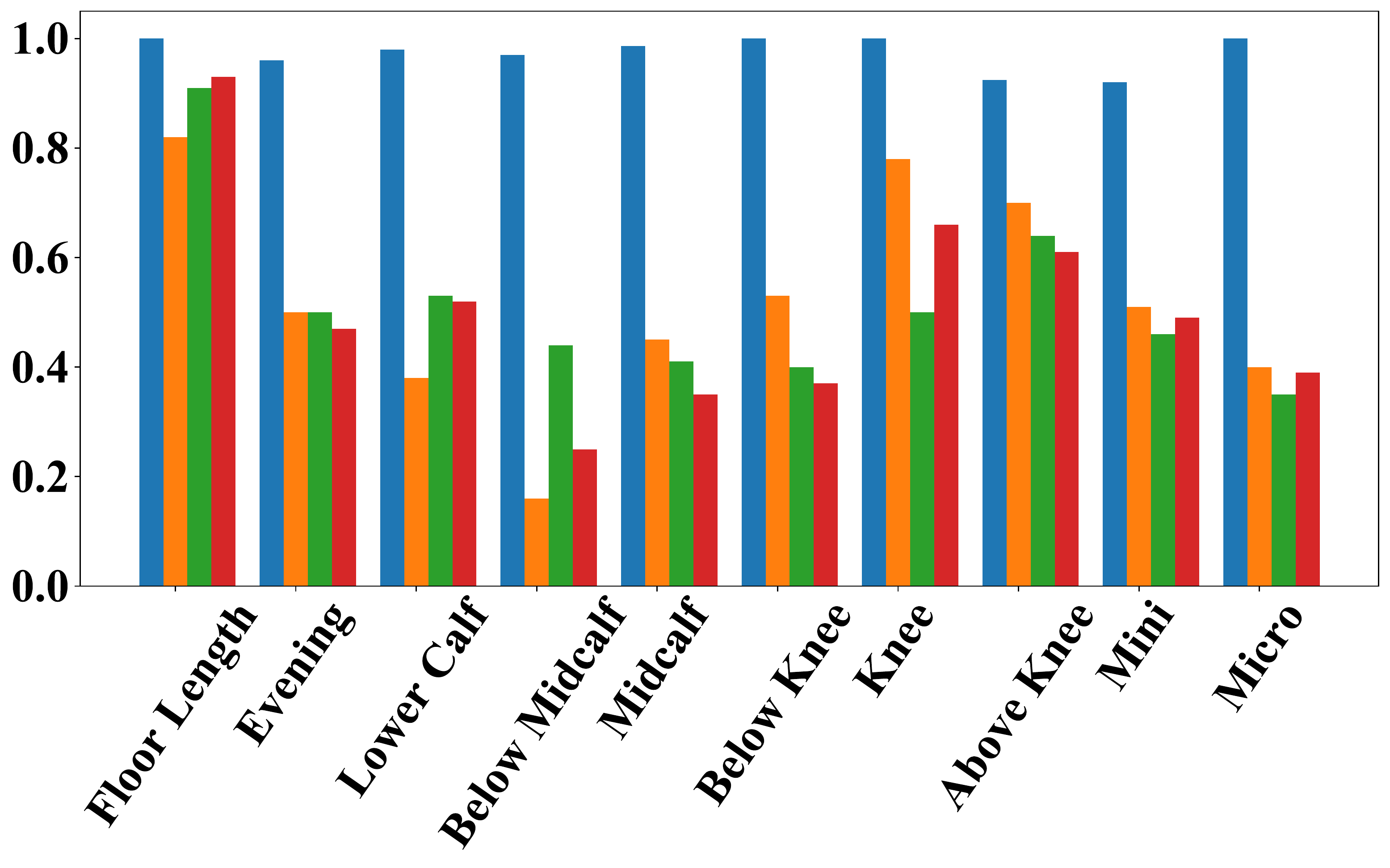}
        \caption{The Precision Results for Predicting Dress Length}
        \label{figure:Dress_Length_Precision}
    \end{subfigure}
    \begin{subfigure}{.5\textwidth}
        \centering
        \includegraphics[width=.8\linewidth]{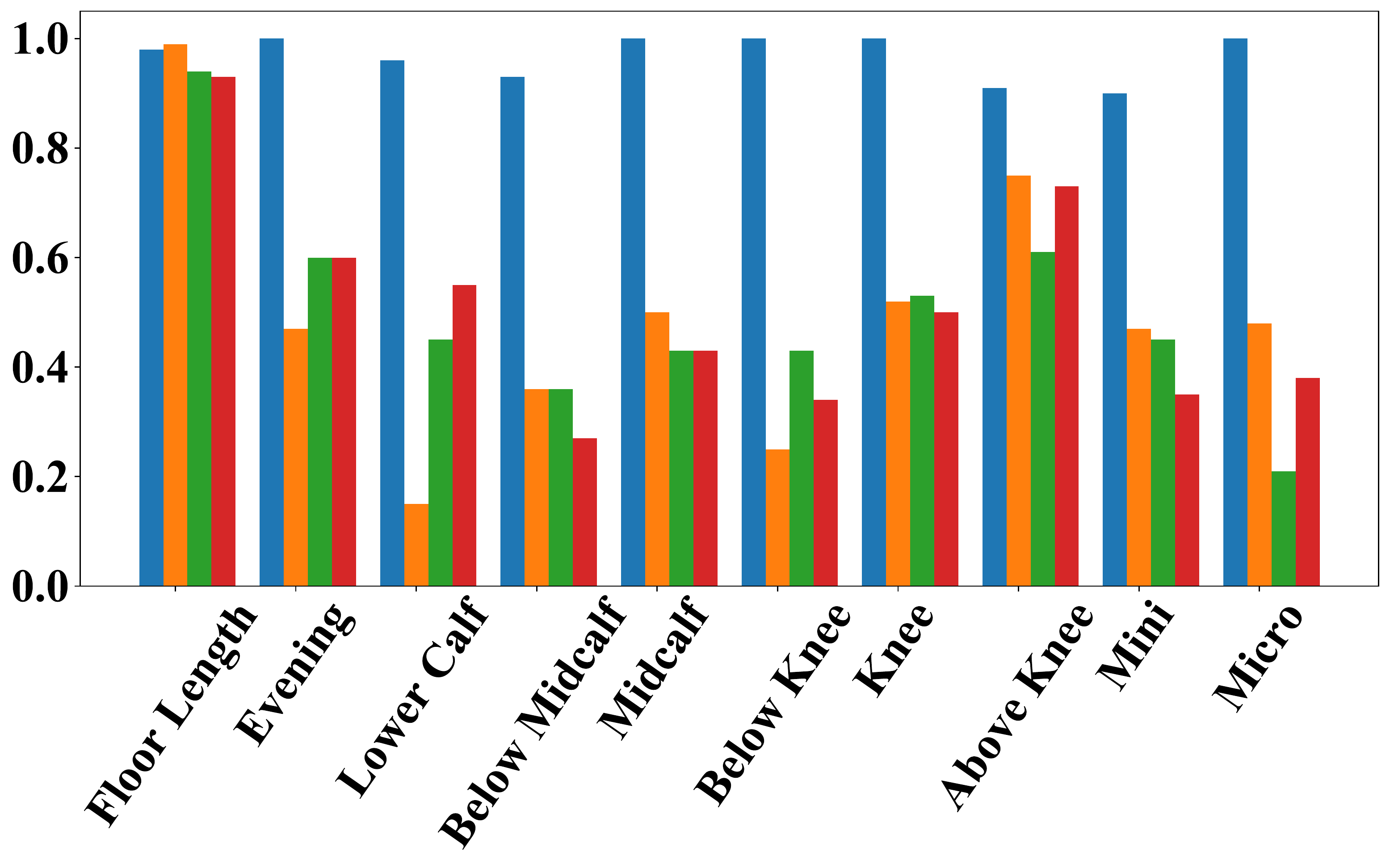}
        \caption{The Recall Results for Predicting Dress Length}
		\label{figure:Dress_Length_Recall}
    \end{subfigure}
    
    \begin{subfigure}{.5\textwidth}
        \centering
        \includegraphics[width=.8\linewidth]{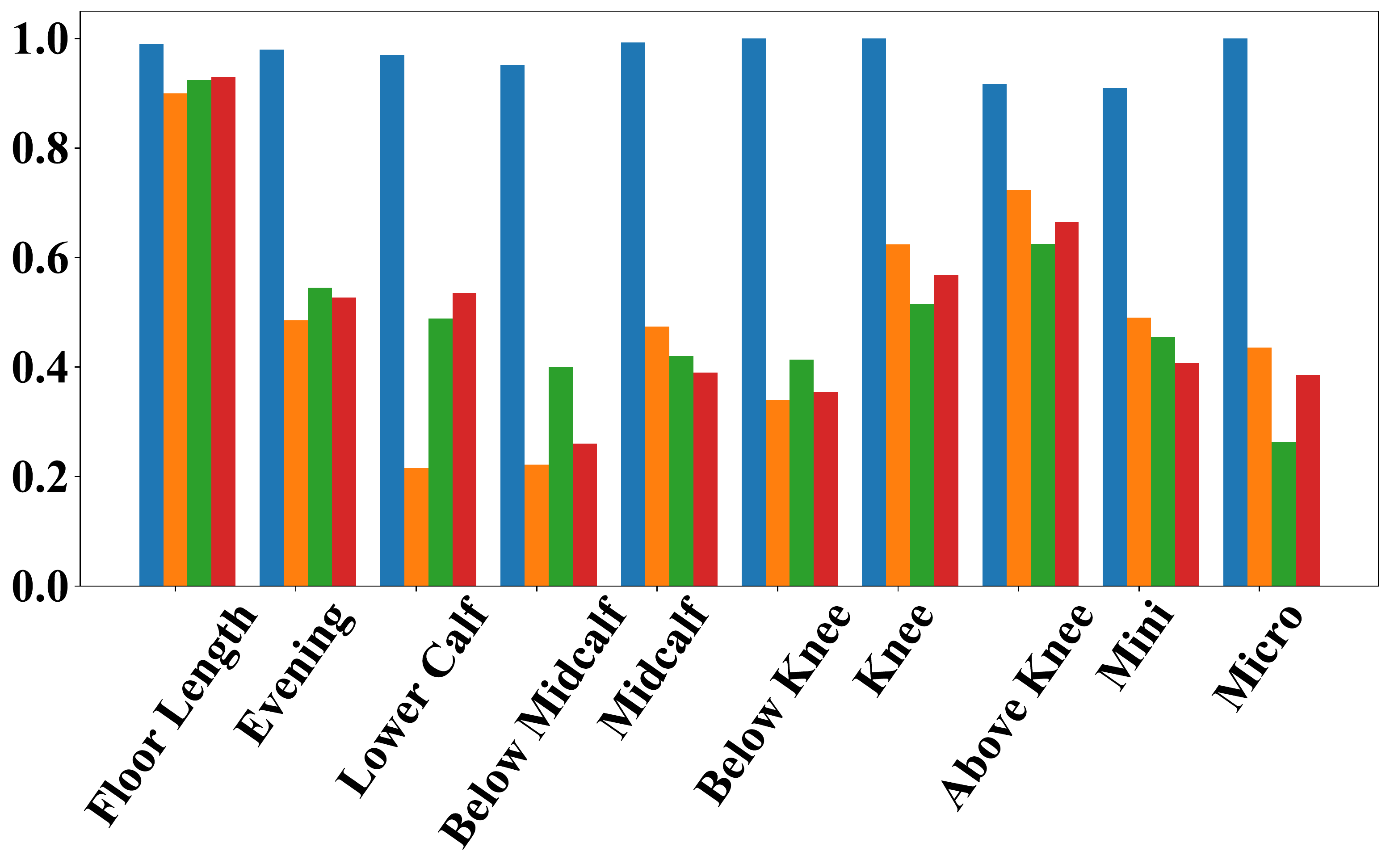}
        \caption{The F1-Score Results for Predicting Dress Length}
		\label{figure:Dress_Length_F1Score}
    \end{subfigure}
    \begin{subfigure}{.5\textwidth}
        \centering
        \includegraphics[width=.8\linewidth]{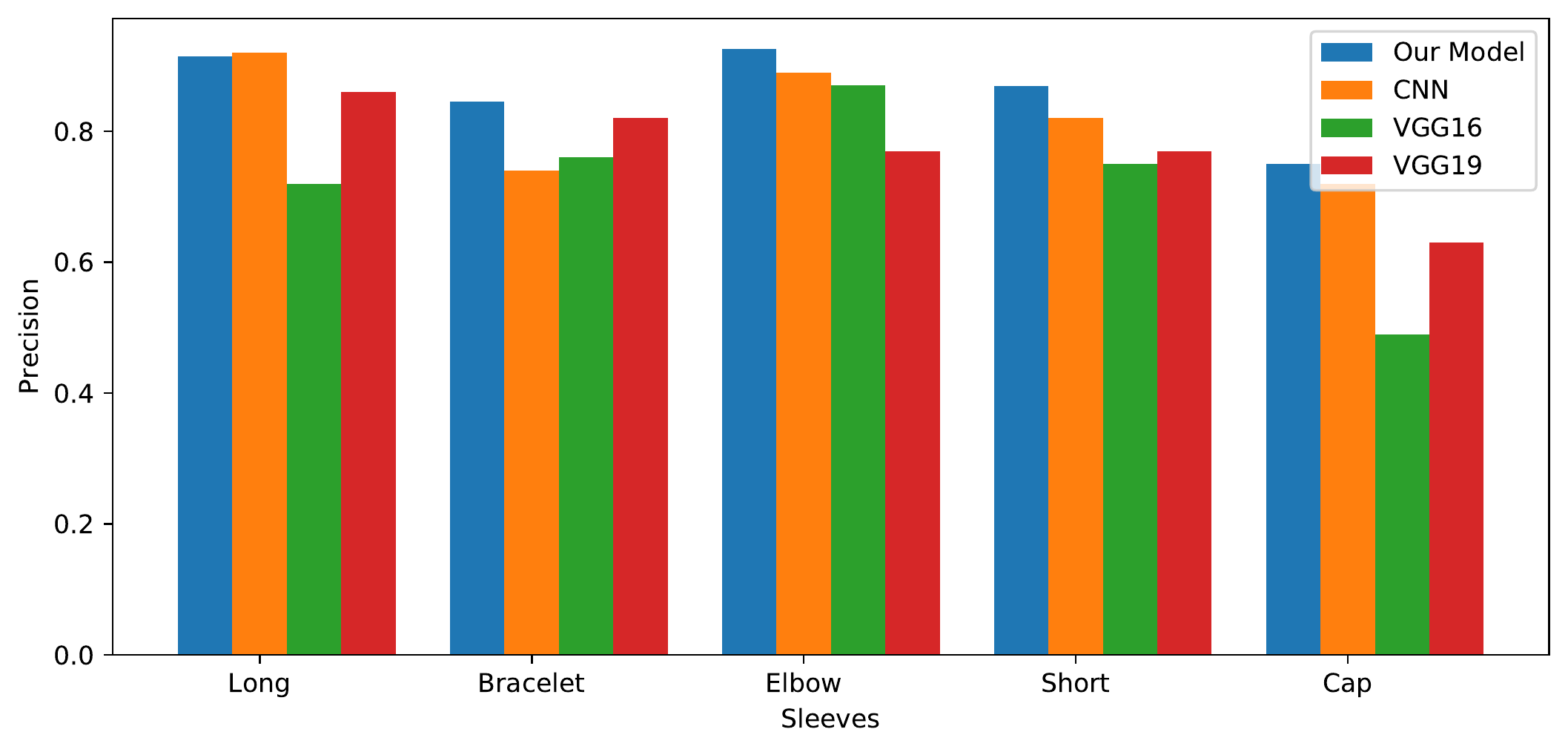}
        \caption{The Precision Results for Predicting Dress Sleeves Type}
		\label{figure:Sleeves_type_Precision}
    \end{subfigure}
    
    \begin{subfigure}{.5\textwidth}
        \centering
        \includegraphics[width=.8\linewidth]{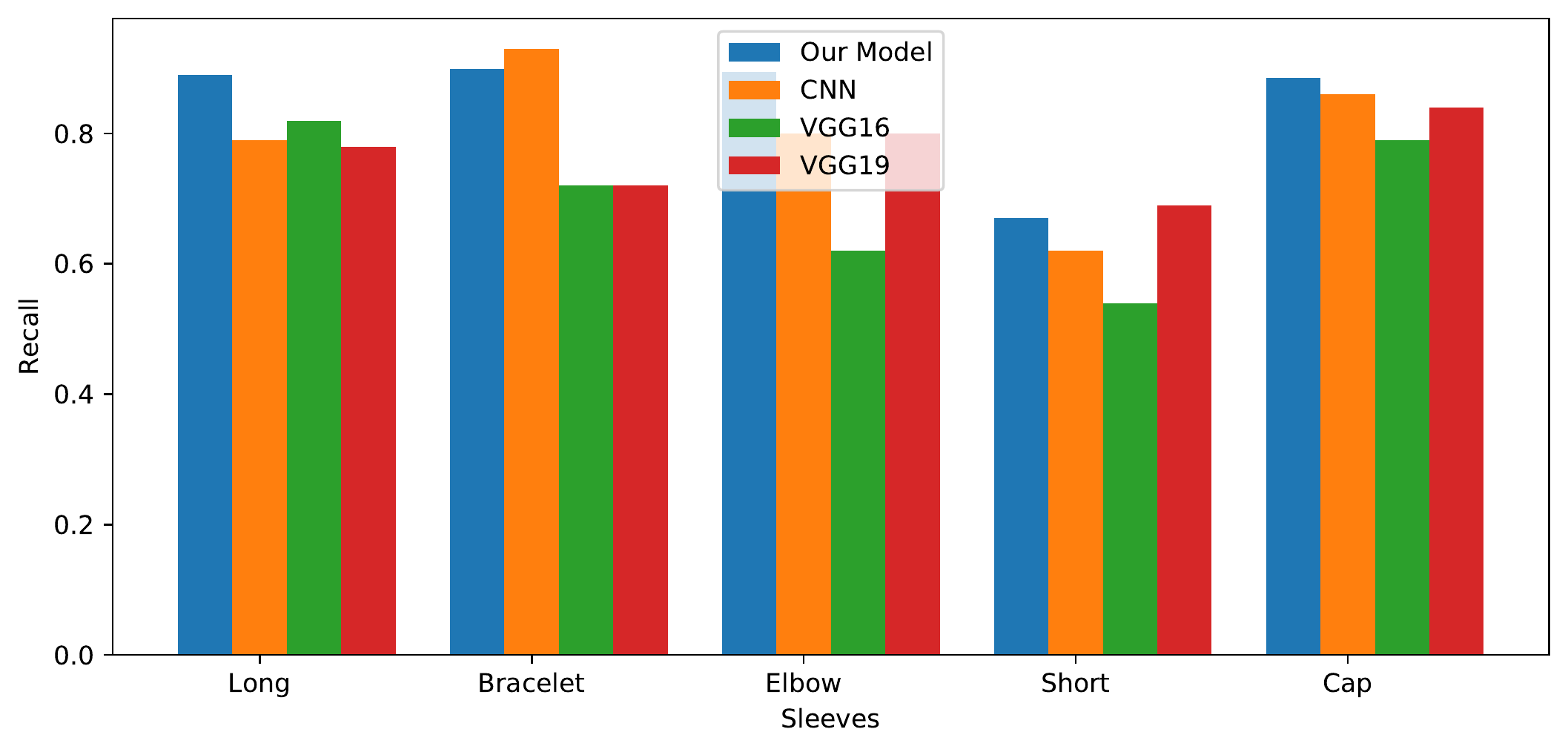}
       \caption{The Recall Results for Predicting Dress Sleeves Type}
		\label{figure:Sleeves_type_Recall}
    \end{subfigure}
    \begin{subfigure}{.5\textwidth}
        \centering
        \includegraphics[width=.8\linewidth]{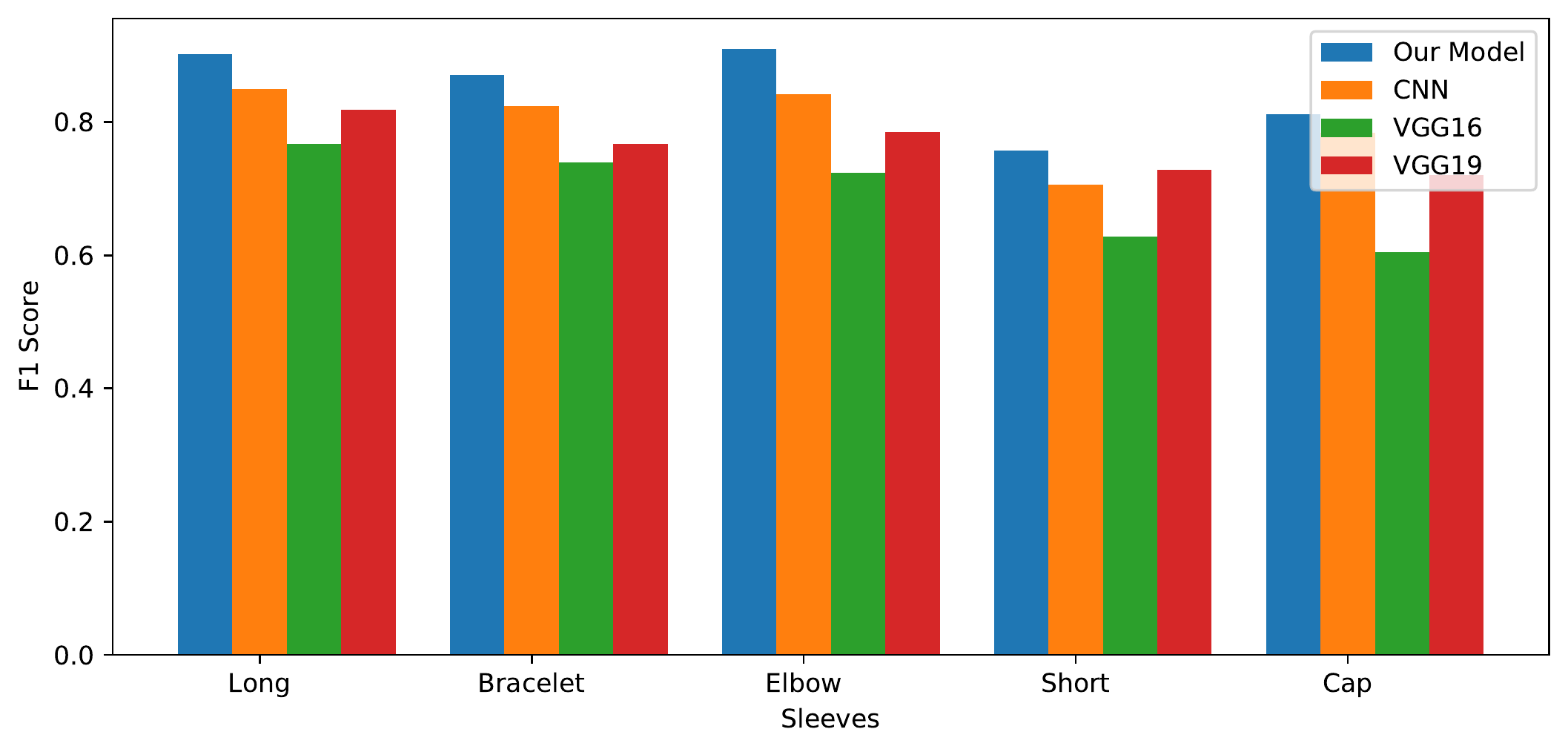}
       \caption{The F1-Score Results for Predicting Dress Sleeves Type}
		\label{figure:Sleeves_type_F1Score}
    \end{subfigure}
    
    \begin{subfigure}{.5\textwidth}
        \centering
        \includegraphics[width=.8\linewidth]{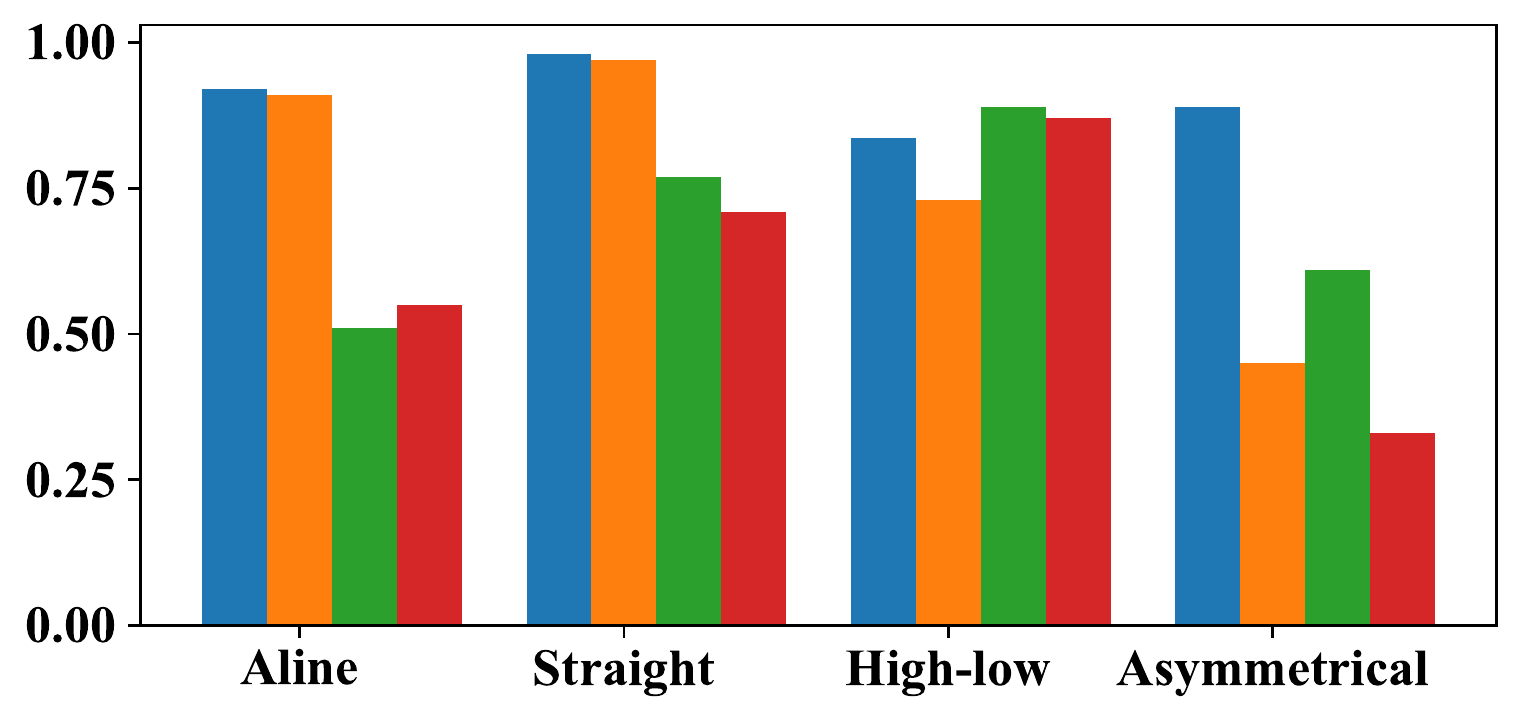}
        \caption{The Precision Results for Predicting Dress Hem Type}
		\label{figure:Hem_type_Precision}
    \end{subfigure}
    \begin{subfigure}{.5\textwidth}
        \centering
        \includegraphics[width=.8\linewidth]{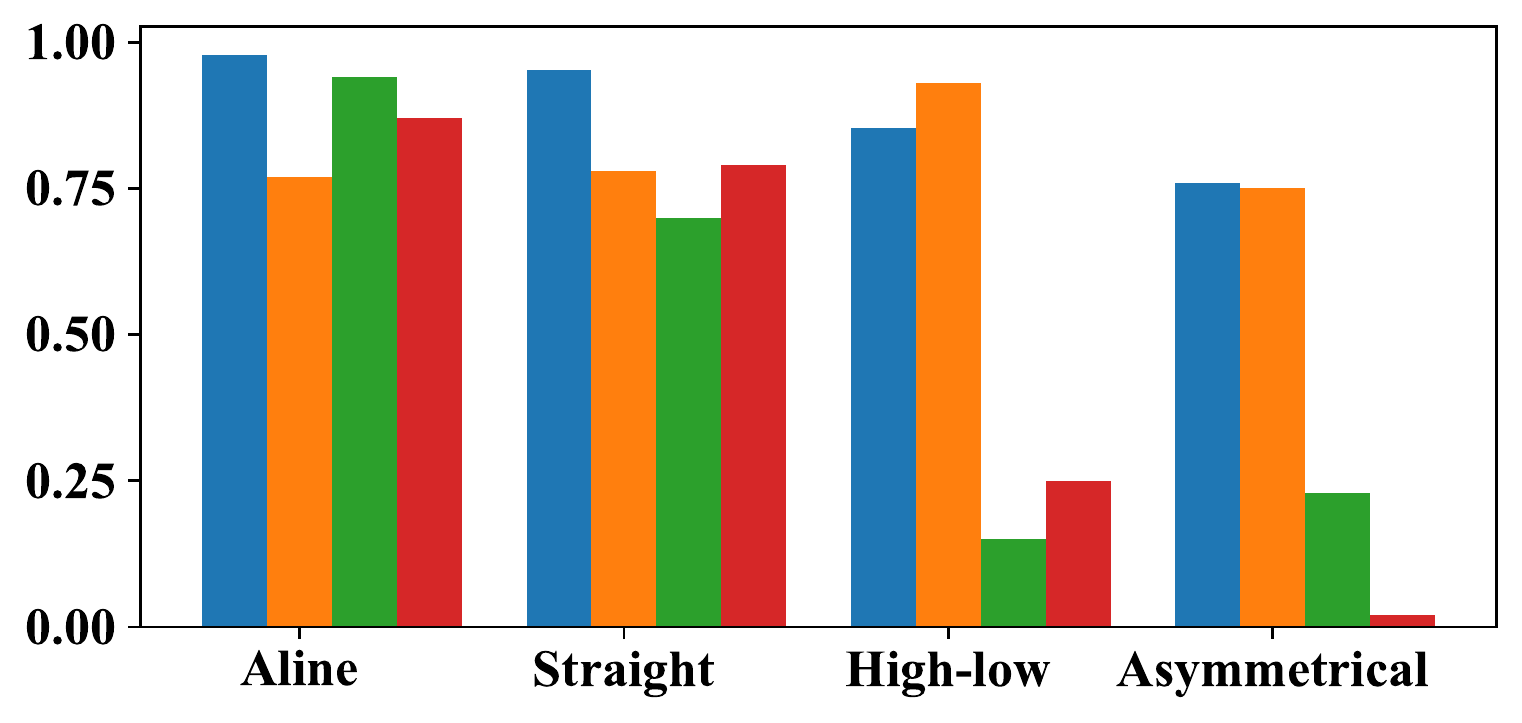}
        \caption{The Recall Results for Predicting Dress Hem Type}
		\label{figure:Hem_type_Recall}
    \end{subfigure}
    
    \begin{subfigure}{.5\textwidth}
        \centering
        \includegraphics[width=.8\linewidth]{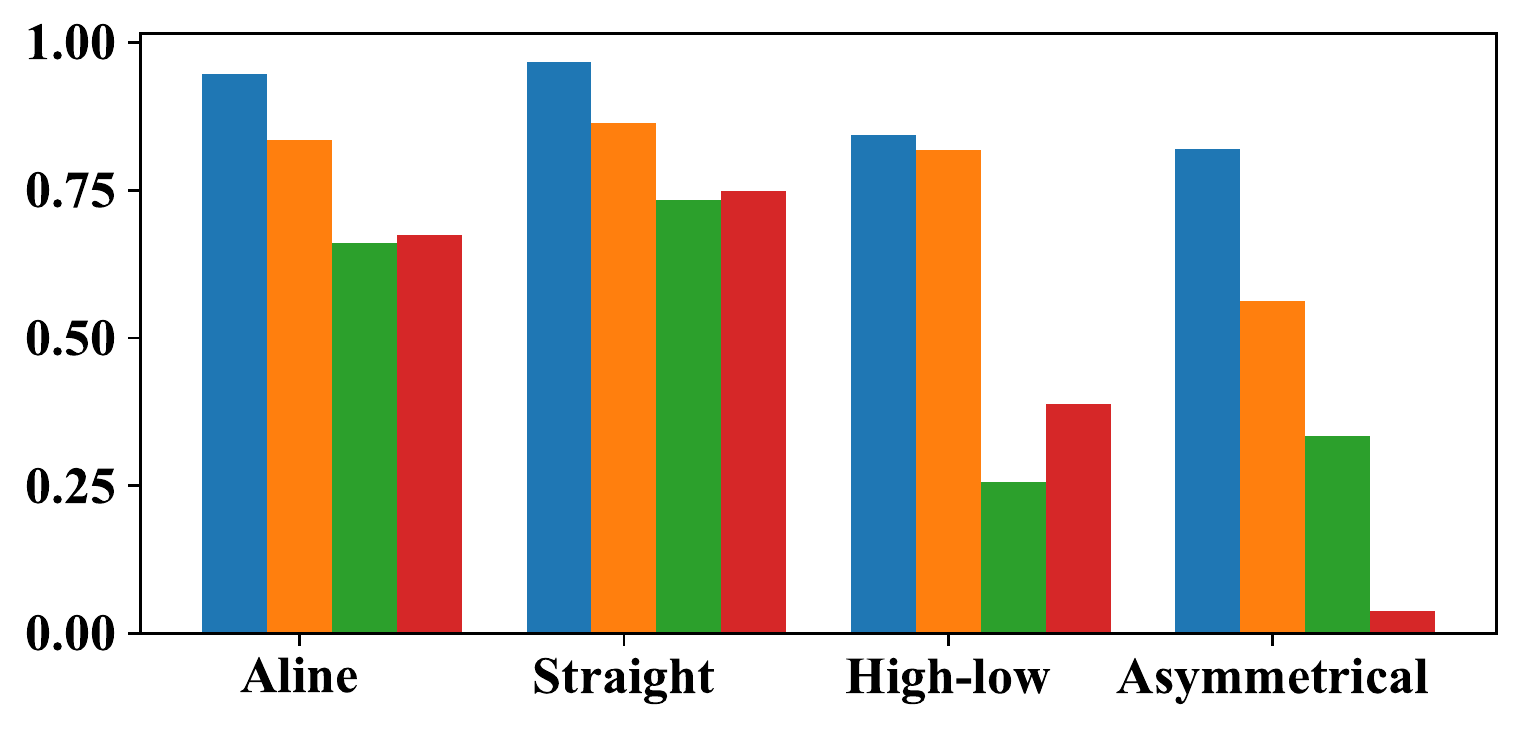}
       \caption{The F1-Score Results for Predicting Dress Hem Type}
		\label{figure:Hem_type_F1Score}
    \end{subfigure}
    \begin{subfigure}{.5\textwidth}
        \centering
        \includegraphics[width=.3\linewidth]{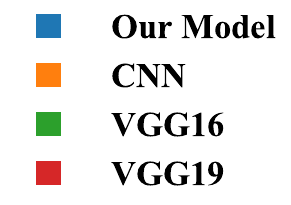}
       \caption{Legend For All Plots}
		\label{figure:legend}
    \end{subfigure}
    \caption{Experimental Results}
\end{figure*}

Furthermore, the plots in Figure 4 emphasize our model's superiority in classification performance to other models. Figures ~\ref{figure:Dress_Length_Precision} to~\ref{figure:Dress_Length_F1Score} represent the performance on hem length data. These plots depict that our proposed model has the highest precision and f1 score over other models.
The sleeve length plots are shown in Figures~\ref{figure:Sleeves_type_Precision} to~\ref{figure:Sleeves_type_F1Score}. Though CNN has the highest recall on bracelet sleeve, our model's average performance scores are far higher than CNN and VGG scores. 
It is evident from the hem type plots in  Figures~\ref{figure:Hem_type_Precision} to~\ref{figure:Hem_type_F1Score} that our proposed scheme achieved prominent classification results when compared against CNN, VGG16 and VGG19 in terms of precision, recall and f1 score metrics


\begin{table}
\centering
\caption{Classification report of our model with metrics precision, recall, and f1-score }
\label{tab:Results_our model}
    \begin{tabular}{ |c|c|c|c|c| } 
     \hline
     \textbf{Class} & \textbf{Attribute} & \textbf{Precision} & \textbf{Recall} & \textbf{F1 Score}\\ \hline \hline
     \multirow{10}{*}{Hem Length} & Floor Length  & 1  & 0.98 & 0.99 \\ \cline{2-5}
     & Evening        & 0.96  & 1       & 0.98 \\ \cline{2-5}
     & Lower Calf     & 0.98  & 0.96    & 0.97  \\ \cline{2-5}
     & Below Midcalf  & 0.976 & 0.93    & 0.952  \\ \cline{2-5}
     & Midcalf        & 0.986 & 1       & 0.993 \\ \cline{2-5}
     & Below Knee     & 1     & 1       & 1  \\ \cline{2-5}
     & Knee           & 1     & 1       & 1\\ \cline{2-5}
     & Above Knee     & 0.924 & 0.91    & 0.917 \\ \cline{2-5}
     & Mini           & 0.92  & 0.90    & 0.91 \\ \cline{2-5}
     & Micro          & 1     & 1       & 1 \\
     \hline
    
     \multirow{5}{*}{Sleeve Length} & Long  & 0.915  &  0.89 &	0.902 \\ \cline{2-5}
     & Bracelet & 0.845  &  0.899 &	0.871  \\ \cline{2-5}
     & Elbow    & 0.926  &  0.895 &	0.91  \\ \cline{2-5}
     & Short    & 0.869  &  0.67  &	0.757 \\ \cline{2-5}
     & Cap      & 0.75   &  0.885 &	0.812 \\ 
     \hline
     \multirow{4}{*}{Hem Type} & Aline  &  0.917 &	0.978 & 	0.947\\  \cline{2-5}
     & Straight      &   0.981 &	0.952 &	0.967 \\ \cline{2-5}
     & High-low      &   0.836 &	0.853 &	0.844 \\ \cline{2-5}
     & Asymmetrical  &   0.89  &	0.76  &	0.82    \\
     \hline
    \end{tabular}
\end{table}
\begin{table}
\centering
\caption{Classification report of CNN with metrics precision, recall, and f1-score }
\label{tab:Results_CNN model}
    \begin{tabular}{ |c|c|c|c|c| } 
     \hline
     \textbf{Class} & \textbf{Attribute} & \textbf{Precision} & \textbf{Recall} & \textbf{F1 Score}\\ \hline \hline
     \multirow{10}{*}{Hem Length} & Floor Length  & 0.82 &	0.99 &	0.9 \\ \cline{2-5}
     & Evening        & 0.5	 &  0.47 &	0.485 \\ \cline{2-5}
     & Lower Calf     & 0.38 &	0.15 &	0.215  \\ \cline{2-5}
     & Below Midcalf  & 0.16 &	0.36 &	0.222  \\ \cline{2-5}
     & Midcalf        & 0.45 &	0.5	 &  0.474 \\ \cline{2-5}
     & Below Knee     & 0.53 &	0.25 &	0.34  \\ \cline{2-5}
     & Knee           & 0.78 &	0.52 &	0.624\\ \cline{2-5}
     & Above Knee     & 0.7	 &  0.75 &	0.724 \\ \cline{2-5}
     & Mini           & 0.51 &	0.47 &	0.49 \\ \cline{2-5}
     & Micro          & 0.4	 &  0.48 &	0.436 \\
     \hline
    
     \multirow{5}{*}{Sleeve Length} & Long  & 0.92 &	0.79 &	0.85 \\ \cline{2-5}
     & Bracelet & 0.74 &	0.93 &	0.824 \\ \cline{2-5}
     & Elbow    & 0.89 &	0.8  &	0.842  \\ \cline{2-5}
     & Short    & 0.82 &	0.62 &	0.706 \\ \cline{2-5}
     & Cap      & 0.72 &	0.86 &	0.784 \\ 
     \hline
     \multirow{4}{*}{Hem Type} & Aline  &  0.91 &	0.77 &	0.834\\  \cline{2-5}
     & Straight      &   0.97 &	0.78 &	0.864 \\ \cline{2-5}
     & High-low      &   0.73 &	0.93 &	0.818 \\ \cline{2-5}
     & Asymmetrical  &   0.45 &	0.75 &	0.562    \\
     \hline
    \end{tabular}
\end{table}

\begin{table}
\centering
\caption{Classification report of VGG16 with metrics precision, recall, and f1-score }
\label{tab:Results_VGG16 model}
    \begin{tabular}{ |c|c|c|c|c| } 
     \hline
     \textbf{Class} & \textbf{Attribute} & \textbf{Precision} & \textbf{Recall} & \textbf{F1 Score}\\ \hline \hline
     \multirow{10}{*}{Hem Length} & Floor Length  & 0.91 &	0.94 &	0.924 \\ \cline{2-5}
     & Evening        & 0.5  &	0.6  &	0.545 \\ \cline{2-5}
     & Lower Calf     & 0.53 &	0.45 &	0.489  \\ \cline{2-5}
     & Below Midcalf  & 0.44 &	0.36 &	0.4  \\ \cline{2-5}
     & Midcalf        & 0.41 &	0.43 &	0.42 \\ \cline{2-5}
     & Below Knee     & 0.4  &  0.43 &	0.414 \\ \cline{2-5}
     & Knee           & 0.5  &	0.53 &	0.515\\ \cline{2-5}
     & Above Knee     & 0.64 &	0.61 &	0.625 \\ \cline{2-5}
     & Mini           & 0.46 &	0.45 &	0.455 \\ \cline{2-5}
     & Micro          & 0.35 &	0.21 &	0.263 \\
     \hline
    
     \multirow{5}{*}{Sleeve Length} & Long  & 0.72 &	0.82 &	0.767 \\ \cline{2-5}
     & Bracelet & 0.76 &	0.72 &	0.739 \\ \cline{2-5}
     & Elbow    & 0.87 &	0.62 &	0.724  \\ \cline{2-5}
     & Short    & 0.75 &	0.54 &	0.628 \\ \cline{2-5}
     & Cap      & 0.49 & 	0.79 &	0.605 \\ 
     \hline
     \multirow{4}{*}{Hem Type} & Aline  &  0.51 &	0.94 &	0.661 \\  \cline{2-5}
     & Straight      &   0.77 &	0.7  &	0.733 \\ \cline{2-5}
     & High-low      &   0.89 &	0.15 &	0.257 \\ \cline{2-5}
     & Asymmetrical  &   0.61 &	0.23 &	0.334   \\
     \hline
    \end{tabular}
\end{table}
\begin{table}
\centering
\caption{Classification report of VGG19 with metrics precision, recall, and f1-score }
\label{tab:Results_VGG19 model}
    \begin{tabular}{ |c|c|c|c|c| } 
     \hline
     \textbf{Class} & \textbf{Attribute} & \textbf{Precision} & \textbf{Recall} & \textbf{F1 Score}\\ \hline \hline
     \multirow{10}{*}{Hem Length} & Floor Length  & 0.93 &	0.93 &	0.93 \\ \cline{2-5}
     & Evening        & 0.47 &	0.6	 &  0.527 \\ \cline{2-5}
     & Lower Calf     & 0.52 &	0.55 &	0.535  \\ \cline{2-5}
     & Below Midcalf  & 0.25 &	0.27 &	0.26  \\ \cline{2-5}
     & Midcalf        & 0.35 &	0.43 &	0.39 \\ \cline{2-5}
     & Below Knee     & 0.37 &	0.34 &	0.354 \\ \cline{2-5}
     & Knee           & 0.66 &	0.5  &	0.569 \\ \cline{2-5}
     & Above Knee     & 0.61 &	0.73 &	0.665 \\ \cline{2-5}
     & Mini           & 0.49 &	0.35 &	0.408 \\ \cline{2-5}
     & Micro          & 0.39 &	0.38 &	0.385 \\
     \hline
    
     \multirow{5}{*}{Sleeve Length} & Long  & 0.86 &	0.78 &	0.8187 \\ \cline{2-5}
     & Bracelet & 0.82 &	0.72 &	0.767 \\ \cline{2-5}
     & Elbow    & 0.77 &	0.8	 &  0.785  \\ \cline{2-5}
     & Short    & 0.77 &	0.8	 &  0.785 \\ \cline{2-5}
     & Cap      & 0.63 &	0.84 &	0.72 \\ 
     \hline
     \multirow{4}{*}{Hem Type} & Aline  &  0.55 &	0.87 &	0.674 \\  \cline{2-5}
     & Straight      &   0.71 &	0.79 &	0.748 \\ \cline{2-5}
     & High-low      &   0.87 &	0.25 &	0.388 \\ \cline{2-5}
     & Asymmetrical  &   0.33 &	0.02 &	0.038   \\
     \hline
    \end{tabular}
\end{table}

\section{Conclusion}\label{sec:conclusion}

This paper proposed a novel algorithm for dress attribute classification. Our approach leverages the key points estimation and the bounding box construction on segmented images. The experimental analysis shows promising results of our model on the three localized categories against CNN, VGG16 and VGG19 models. The benefit of our approach is three-fold. First, our model is computationally efficient and thus easily applicable by a broad range of online retailers, unlike CNN-based approaches whose time complexity depends on level of architecture complexity. Second, our approach produces robust results even with small datasets whereas CNN based models require a huge amount of data to reduce over fitting. A third benefit is the feature extractors are independent of one another. They can be refined or replaced as better approaches emerge without adversely affecting the performance of other extractors. With that in mind, our comparison of algorithmic approaches to classification versus CNN and VGG was less about demonstrating the superiority of our approach and more about being certain that there was not a better way to implement a specific feature extractor.
In the future study, we will extend the proposed scheme to classify additional categories of dress, e.g., necklines (U-neck, Square, V-neck, etc), patterns (striped, dotted, floral, etc.), textures (denim, cotton, leather, etc.), and so forth.


%

\bibliographystyle{IEEEtran}
\bibliography{bare_jrnl_compsoc}



%








\end{document}